\documentclass[10pt,twocolumn,letterpaper]{article}

\usepackage{cvpr}
\usepackage{times}
\usepackage{epsfig}
\usepackage{graphicx}
\usepackage{amsmath}
\usepackage{amssymb}
\usepackage{nccmath}
\usepackage{subcaption}
\usepackage{adjustbox}
\usepackage{enumitem}
\usepackage{pbox}
\usepackage{makecell}
\usepackage{cellspace}
\usepackage{booktabs}
\usepackage{float}
\usepackage{multicol}

\usepackage{xcolor}

\usepackage[pagebackref=true,breaklinks=true,letterpaper=true,colorlinks,bookmarks=false]{hyperref}

\newcommand{\ModelName}{ViDeNN}

\cvprfinalcopy 

\def\cvprPaperID{****} 

\ifcvprfinal\fi

\begin{document}
\setlength{\abovedisplayskip}{6pt}
\setlength{\belowdisplayskip}{6pt}
\title{ViDeNN: Deep Blind Video Denoising }

\author{Michele Claus\\
Computer Vision lab\\
Delft University of Technology\\
{\tt\small claus.michele@hotmail.it}
\and
Jan van Gemert\\
Computer Vision lab\\
Delft University of Technology\\
{\tt\small http://jvgemert.github.io/}
}

\maketitle
\begin{abstract}
We propose \ModelName: a CNN for Video Denoising without prior knowledge on the noise distribution (blind denoising). The CNN architecture uses a combination of spatial and temporal filtering, learning to spatially denoise the frames first and at the same time how to combine their temporal information, handling objects motion, brightness changes, low-light conditions and temporal inconsistencies. We demonstrate the importance of the data used for CNNs training, creating for this purpose a specific dataset for low-light conditions. We test \ModelName~on common benchmarks and on self-collected data, achieving good results comparable with the state-of-the-art. 
\end{abstract}

\section{Introduction}
Image and video denoising aims to obtain the original signal $X$ from available noisy observations $Y$. Noise influences the perceived visual quality, but also segmentation~\cite{MRI} and compression~\cite{JPEG2000} making denoising an important step. With $X$ as the original signal, $N$ as the noise and $Y$ as the available noisy observation, the noise degradation model can be described as $Y = X + N$, for an additive type of noise. In low-light conditions, noise is signal dependent and more sensitive in dark regions, modeled as $Y = H(X) + N$, with $H$ as the degradation function.

\begin{figure}
\begin{center}
\includegraphics[width=0.36\textwidth]{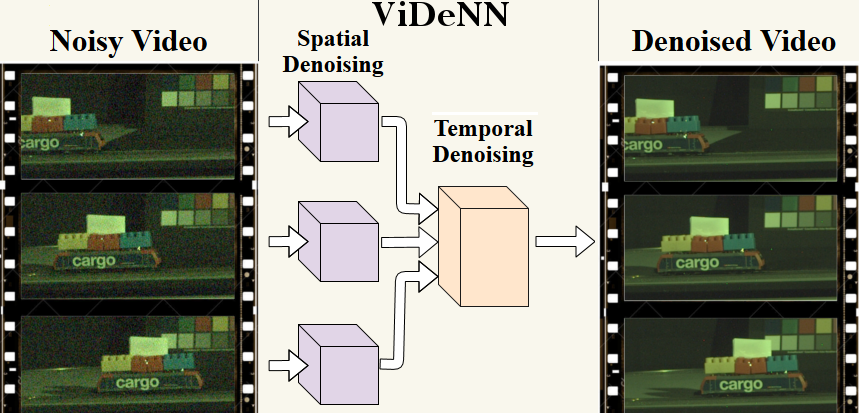}
\end{center}
\setlength{\belowcaptionskip}{-10pt}
\caption{\ModelName~approach to Video Denoising: combining two networks performing first Single Frame Spatial Denoising and subsequently Temporal Denoising over a window of three frames, all in a single feed-forward process.}
\label{fig:figure1}
\end{figure}

Imaging noise is due to thermal effects, sensor imperfections or low-light. Hand tuning multiple filter parameters is fundamental to optimize quality and bandwidth of new cameras for each gain level, taking much time and effort. Here, we automate the denoising procedure with a CNN for flexible and efficient video denoising, capable to blindly remove noise. Having a noise removal algorithm working in ``blind'' conditions is essential in a real-world scenario where color and light conditions can change suddenly, producing a different noise distribution for each frame. 

Solutions based on statistical models include Markov Random Field models \cite{MRF}, gradient models \cite{GRAD}, sparse models \cite{SPAR} and Nonlocal Self-Similarity (NSS) currently used in state-of-the-art techniques such as BM3D \cite{BM3D}, LSSC \cite{LSSC}, NCSR \cite{NCSR} and WNNM \cite{WNNM}. Even though they achieve respectable denoising performance, most of those algorithms have some drawbacks. Firstly, they are generally designed to tackle specific noise models and levels, limiting their usage in blind denoising. Secondly, they involve time-consuming hand-tuned optimization procedures. 

Much work has been done on image denoising while few algorithms have been specifically designed for videos. The key assumption for video denoising is that video frames are strongly correlated. The most basic video denoising technique consists of the temporal average over various subsequent frames. While giving excellent results for steady scenes, it blurs motion, causing artifacts. The VBM4D method \cite{VBM4D} is the state-of-the-art in video denoising. It extends BM3D~\cite{BM3D} single image denoising by the search of similar patches, not only in spatial but also in temporal domain. Searching for similar patches in more frames drastically increases the processing time. 

In this paper we propose \ModelName, illustrated in Fig~\ref{fig:figure1}: a convolutional neural network for blind video denoising, capable to denoise videos without prior knowledge over the noise model and the video content.
For comparison, experiments have been run on publicly available and on self captured videos. The videos, the low-light dataset and the code will be published on the project's GitHub page.\\
The main contributions of our work are: (i) a novel CNN architecture capable to blind denoise videos, combining spatial and temporal information of multiple frames with one single feed-forward process; (ii) Flexibility tests on Additive White Gaussian Noise and real data in low-light condition; (iii) Robustness to motion in challenging situations; (iv) A new low-light dataset for a specific Bosch security camera, with sample pairs of noise-free and noisy images.

\section{Related Work}
\textbf{CNNs for Image Denoising.} From the 2008 CNN image denoising work of Jain and Seung \cite{NIPS2008_3506} there have been huge improvements thanks to more computational power and high quality datasets. In 2012, Burger \etal~\cite{MLP} showed how even a simple Multi Layer Perceptron can obtain comparable results with BM3D \cite{BM3D}, even though a huge dataset was required for training \cite{LABELME}. Recently, in 2016, Zhang \etal~\cite{DNCNN} used residual learning and Batch Normalization \cite{BATCH} for image denoising in their DnCNN architecture. With its simple yet effective architecture, it has shown to be flexible for tasks as blind Gaussian denoising, JPEG deblocking and image inpainting. FFDNet \cite{FFDNET} extends DnCNN by handling an extended range of noise levels and has the ability to remove spatially variant noise. Ulyanov \etal~\cite{DEEPPRIOR} showed how, with their Deep Prior, they can enhance a given image with no prior training data other than the image itself, which can be seen as a ''blind'' denoising. There have been also some works on CNNs directly inspired by BM3D such as \cite{LEFKI,BM3DNET}. In \cite{MEMNET}, Ying \etal propose a deep persistent memory network called MemNet that obtains valid results, introducing a memory block, motivated by the fact that human thoughts are persistent. However, the network structure remains complex and not easily reproducible.
A U-Net variant has been successfully used for image denoising in the work of Xiao-Jiao \etal~\cite{RED30} and in the most recent work on image denoising of Guo \etal~\cite{CBDNET} called CBDNet. With their novel approach, CBDNet reaches extraordinarily results in real world blind image denoising.
The recently proposed Noise2Noise \cite{NOISE2NOISE} model is based on an encoder-decoder structure, obtains almost the same result using only noisy images for training, instead of clean-noisy pairs, which is particularly useful for cases where the ground truth is not available.

\textbf{Video and Deep Neural Networks.}
Video denoising using deep learning is still an under-explored research area. The seminal work of Xinyuan \etal~\cite{RNN}, is currently the only one using neural networks (Recurrent Neural Networks) to address video denoising. We differ by addressing color video denoising, and offer comparable results to the state-of-art. Other video-based tasks addressed using CNNs include Video Frame Enhancement, Interpolation, Deblurring and Super-Resolution, where the key component is how to handle motion and temporal changes. For frame interpolation, Niklaus \etal~\cite{Niklaus18} use a pre-computed optical flow to feed motion information to a frame interpolation CNN. Meyer \etal~\cite{Meyer18} use instead phase based features to describe motion. Caballero \etal~\cite{Caballero} developed a network which estimate the motion by itself for video super resolution. Similarly, in Multi Frame Quality Enhancement (MFQE), Yang \etal~\cite{MFQE} use a Motion Compensation Network and a Quality Enhancement Network, considering three non-consecutive frames for H265 compressed videos. Specifically for video deblurring, Su \etal~\cite{Su17} developed a network called DeBlurNet: a U-Net CNN which takes three frames stacked together as input. Similarly, we also use three stacked frames in our \ModelName. Additionally, we have also investigated variations in the number of input frames.

\textbf{Real World Datasets.}
An image or video denoising algorithm, has to be effective on real world data to be successful. However, it is hard to obtain the ground truth for real pictures, since perfect sensors and channels do not exist. In 2014, Anaya and Barbu, created a dataset for low-light conditions called RENOIR \cite{RENOIR}: they use different exposure times and ISO levels to get noisy and clean images of the same static scene. Similarly, in 2017, Plotz and Roth created a dataset called DND \cite{DND}. In this case, only the noisy samples have been released, whereas the noise free ones are kept undisclosed for benchmarking purposes. Recently, two other related papers have been published. The first, written by Abdelhamed \etal~\cite{SMARTPHONE} concerns the creation of a smartphone image dataset of noisy and clean images, which at the time of writing is not yet publicly available. The second, written by Chen \etal~\cite{DARK}, presents a new CNN based algorithm capable to enhance the quality of low-light raw images. They created a dedicated dataset of two camera types similarly to \cite{RENOIR}.

\begin{figure}
\begin{center}
\includegraphics[width=0.5\textwidth]{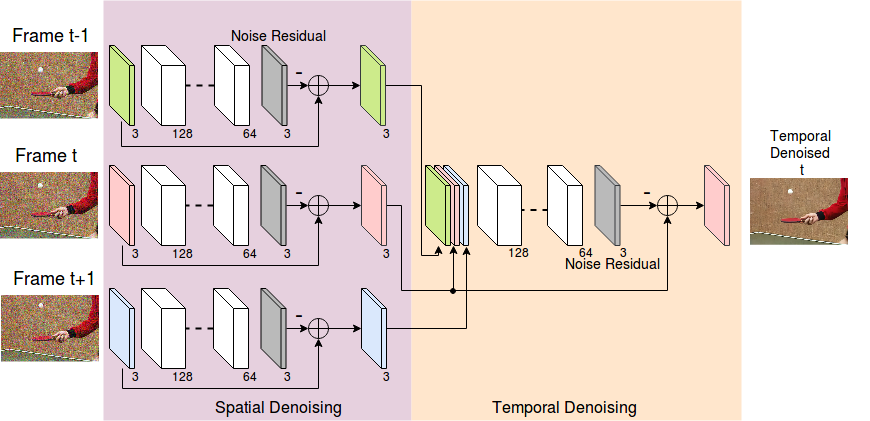}
\end{center}
\setlength{\belowcaptionskip}{-9pt}
\caption{The architecture of the proposed \ModelName~network. Every frame will go through a spatial denoising CNN. The temporal CNN takes as input three spatially denoised frames and outputs the final estimate of the central frame. Both CNNs estimate first the noise residual, \textit{i.e.} the unwanted values noise adds to an image, and then subtracts them from the noisy input ($\oplus$ means addition of the two signals, and ''-'' the negation). \ModelName~is composed only by Convolutional Layers. The number of feature maps is written at the bottom of each layer.}
\label{fig:vidcnn}
\end{figure}

\section{\ModelName: Video DeNoising Net}

\ModelName~has two subnetworks: Spatial and temporal denoising CNN, as illustrated in Fig~\ref{fig:vidcnn}.

\subsection{Spatial Denoising CNN}
For spatial denoising we build on~\cite{DNCNN}, which showed great flexibility tackling multiple degradation types at the same time, and experimented with the same architecture for blind spatial denoising. It is shown, that this architecture can achieve state-of-art results for Gaussian denoising. A first layer of depth 128 helps when the network has to handle different noise models at the same time. The network depth is set to 20 and Batch Normalization (BN) \cite{BATCH} is used. The activation function is ReLU (Rectified Linear Unit). We also investigated the use of Leaky ReLU as activation function, which can be more effective \cite{xu2015empirical}, without improvement over ReLU. Comparison results are provided in the supplementary material.
Our Spatial-CNN uses Residual Learning, which has been firstly introduced in \cite{DNCNN} to tackle image denoising. Instead of forcing the network to output directly the denoised frame, the residual architecture predicts the residual image, which consist in the difference between the original clean image and the noisy observation.
The loss function $L$ is the L2-norm, also known as least squares error (LSE) and is the sum of the square of the differences $S$ between the target value $Y$ and the estimated values $Y_{est}$. In this case the difference $S$ represents the noise residual image estimated by the Spatial-CNN, and is given by $L=\sum\limits_{x}\sum\limits_{y}\Big(\underbrace{Y(x,y)-Y_{est}(x,y)}_\text{Noise Residual}\Big)^2 \label{eq:1}$.

\subsubsection{A Realistic Noise Model}
\begin{figure}
\centering
\begin{subfigure}{.11\textwidth}
  \centering
  \captionsetup{justification=centering}
  \includegraphics[width=.9\linewidth]{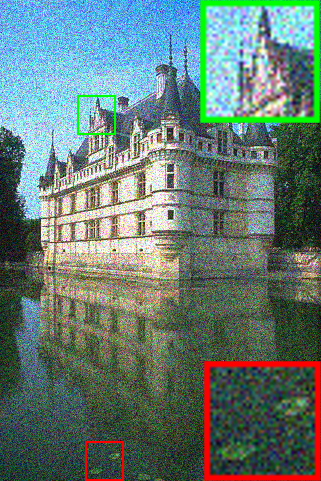}
  \caption{\\Noisy frame\\(18.54/0.5225)}
  \label{fig:nfig1}
\end{subfigure}
\begin{subfigure}{.11\textwidth}
  \centering
  \captionsetup{justification=centering}
  \includegraphics[width=.9\linewidth]{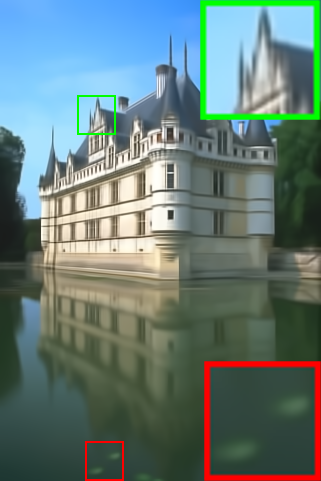}
  \caption{CBM3D\cite{CBM3D}\\(29.26/0.9194)}
  \label{fig:nfig2}
\end{subfigure}
\begin{subfigure}{.11\textwidth}
  \centering
  \captionsetup{justification=centering}
  \includegraphics[width=.9\linewidth]{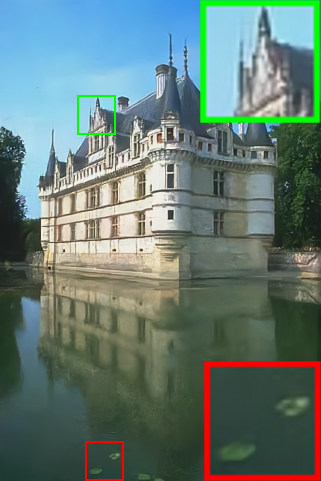}
  \caption{DnCNNB\cite{DNCNN}\\(28.72/0.9355)}
  \label{fig:nfig3}
\end{subfigure}
\begin{subfigure}{.11\textwidth}
  \centering
  \captionsetup{justification=centering}
  \includegraphics[width=.9\linewidth]{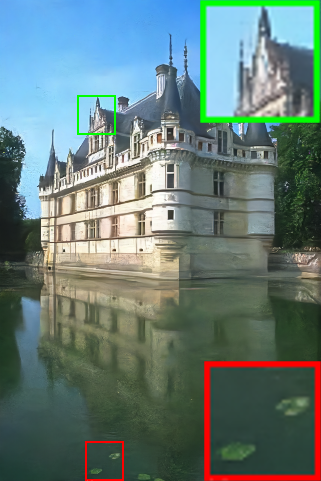}
  \caption{\\Our result\\(30.37/0.9361)}
  \label{fig:nfig5}
\end{subfigure}
\setlength{\belowcaptionskip}{-8pt}
\caption{Comparison of spatial denoising of an image from the CBSD68 dataset corrupted with \ref{eq:2}, with Ag=64 and Dg=4. AWGN based method as CBM3D and DnCNN does not achieve optimal result. The first blurs excessively the image. Using the proper noise model for training leads to a better result. (PSNR [dB]/SSIM)}
\label{fig:fig2}
\end{figure}

The denoising performance of a spatial denoising CNN depends greatly on the training data. Real noise distribution differs from Gaussian, since it is not purely additive but it contains a \textit{signal dependent} part. For this reason, CNN models trained only on Additive White Gaussian Noise (AWGN) fail to denoise real world images \cite{CBDNET}.
Our goal is to achieve a good balance between performance and flexibility, training a single network for multiple noise models. As shown in Table \ref{tab:tab1}, our Spatial-CNN can handle blind Gaussian denoising: we will further investigate its generalization capabilities, introducing a signal dependent noise model. This specific noise model, in equation \ref{eq:2}, is composed by two main contributions, the Photon Shot Noise (PSN) and the Read Noise.
The PSN is the main noise source in low-light condition, where $N_{sat}$ accounts the saturation number of electrons. The Read Noise is mainly due to the quantization process in the Analog to Digital Converter (ADC), used to transform the analog light signal into a digital image. $CT1_n$ represents the normalized value of the noise contribution due to the Analog Gain, whereas $CT2_n$ represents the additive normalized part. The realistic noise model is
\begin{gather}
M=\sqrt[]{\underbrace{\frac{Ag*Dg}{N_{sat}*s}}_\text{PSN}+\underbrace{Dg^2*(Ag*CT1_n+CT2_n)^2}_\text{Read Noise}}, \label{eq:2}\\
\text{Noisy Image} = s + \mathcal{N}(0,1)*M(s), \label{eq:3}
\end{gather}
where the relevant terms for the considered Sony sensor are: $Ag$ (Analog Gain), in range [0,64], $Dg$ (Digital Gain), in range [0,32] and $s$, the image that will be degraded. The remaining values are $CT1_n$=$1.25^{-4}$, $CT2_n$=$1.11^{-4}$ and $N_{sat}$=$7489$.
The noisy image is generated by multiplying observations of a normal distribution $\mathcal{N}(0,1)$ with the same shape of the reference image $s$, with the Noise Model $M$ in equation \ref{eq:3}.
In Figure \ref{fig:fig2} we illustrate that AWGN-based algorithms such as CBM3D and DnCNN do not generalize to realistic noise. CBM3D, in its blind version, i.e. with the supposed AWGN standard deviation $\sigma$ set to 50, over-smooths the image, getting a low SSIM (Structural Similarity, the higher the better) score, whereas DnCNN preserves more structure.
Our result shows that, to better denoise real world images, a realistic noise model has to be used for the training set.

\subsection{Temp3-CNN: Temporal Denoising CNN}
The temporal denoising part of \ModelName~is similar in structure to the spatial one, having the same number of layers and feature maps. However, it stacks frames together as input. Following other work~\cite{Caballero,MFQE,Su17} we stack 3 frames, which is efficient, while our preliminary results show no improvements for stacking more than 3 frames. Considering a frame with dimensions $w{\times}h{\times}c$, the new input will have dimension of $w{\times}h{\times}3c$. 
Similar to the Spatial-CNN, the temporal denoising also uses residual learning and will estimate the noise residual image of the central input frame, combining the information of other frames allowing it to learn temporal inconsistencies. 

\section{Experiments}
In this section we present the dataset creation and the training/validation, performing insightful experiments.
\subsection{Low-Light Dataset Creation}
An image denoising dataset has pairs of clean and noisy images. For realistic low-light conditions, creating pairs of noisy and clean images is challenging and the publicly available data is scarce. We used Renoir \cite{RENOIR} and our self-collected dataset.
Renoir~\cite{RENOIR} proposes to use two different ISO values and exposure times to get reference and distorted images, demanding many camera settings and parameters. We use a simpler process: grabbing many noisy images of a static scene and then simply averaging to get an estimated ground truth. We used a Bosch Autodome IP 5000 IR, a security camera capable to record raw images, \textit{i.e.} without any type of processing. The setting involved a static scene and a light source with color temperature $3460K$, which has variable intensity between 0 and 255. 
We varied the light intensity in 12 steps, from the lowest acceptable light condition of value 46, below of which the camera showed noise only, up to the maximum with value 255. For every different light intensity, we recorded 200 raw images in a row. Additionally, we recorded six test video sequences with the stop-motion technique in different light conditions, consisting in three or four frames with moving objects or light changes: for each frame we recorded 200 images, which results in a total of 4200 images.

\begin{figure}
\centering
\begin{subfigure}{.18\textwidth}
  \centering
  \captionsetup{justification=centering}
  \includegraphics[width=\textwidth]{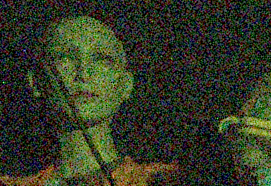}
  \caption{Low-light Noisy\\Image}
  \label{fig:bosch1}
\end{subfigure}
\begin{subfigure}{.18\textwidth}
  \centering
  \captionsetup{justification=centering}
  \adjincludegraphics[width=\textwidth]{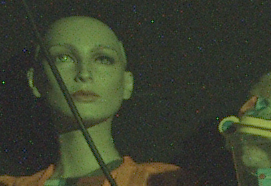}
  \caption{Reference Ground Truth}
  \label{fig:bosch2}
\end{subfigure}
\setlength{\belowcaptionskip}{-10pt}
\caption{Sample detail of noisy-clean image pairs of our own low-light dataset, collected with a Bosch Autodome IP 5000 IR security camera. The ground truth is obtained averaging 200 raw images collected in the same light conditions.}
\label{fig:bosch4}
\end{figure}

\subsection{Spatial CNN Training}
The training is divided in two parts: (i) we train the spatial denoising CNN and (2) after convergence, we train the temporal denoising CNN.

Our ideal model has to tackle multiple degradation types at the same time, such as AWGN and real noise model \ref{eq:3} including low-light conditions. During the training phase, our neural network will learn how to estimate the residual noise content of the input noisy image, using the clean one as reference. Therefore, we require couples of clean and noisy images. which are easily created for AWGN and the real noise model in equation \ref{eq:3}. 

We use the Waterloo Exploration Dataset \cite{WATERLOO}, containing 4,744 clean images. The amount of available images helps greatly to generalize and allows us to keep a good part of it for testing. The dataset is randomly divided in two parts, 70\% for training and 30\% for testing. Half of the images are being added with AWGN with $\sigma$=[0,55]. The second half are processed with equation \ref{eq:3} which is the realistic noise model, with Analog Gain Ag=[0,64] and Digital Gain Dg=[0,32]. 

Following \cite{DNCNN}, the network is trained with $50 \times 50 \times 3$ patches. We obtained 120,000 patches from the Waterloo training set, containing AWGN and real noise type, using data augmentation such as rotating, flipping and mirroring. For low-light conditions, we used five noisy images for each light level from our own training dataset, obtaining 60 pairs of noisy-clean images for training. The patches extracted are 80,000. From the Renoir dataset, we used the subset T3 and randomly cropped 40,000 patches. For low-light testing, we will use 5 images from our camera of a different scene, not present in the training set, and part of the Renoir T3 set. We trained for 100 epochs, using a batch of 128 and Adam Optimizer \cite{ADAM} with a learning rate of $10^{-3}$ for the first 20 epochs and $10^{-4}$ for the latest 80. 
\subsection{Validation of static Image Denoising}
We compared blind Gaussian denoising with the original implementation of DnCNN, on which ours is based. From our test in Table \ref{tab:tab1} on the BSD68 test set, we notice how the result of our blind model and the one proposed by the paper \cite{DNCNN} are comparable.

\begin{table}
\begin{center}
\begin{adjustbox}{width=0.48\textwidth,center}
\begin{tabular}{@{}lcccccc@{}}
\toprule
 & $\sigma=5$ & $\sigma=10$ & $\sigma=15$ & $\sigma=25$ & $\sigma=35$ & $\sigma=50$ \\
\midrule
Spatial-CNN*& 39.73 & 35.92 & 33.66 & 30.99 & 29.34 & 27.63 \\
DnCNN-B* \cite{DNCNN}& 39.79 & 35.87 & 33.57 & 30.69 & 28.74 & 26.53 \\
DnCNN-B \cite{DNCNN}& 40.62 & 36.14 & 33.88 & 31.22 & 29.57 & 27.91 \\
\bottomrule
\end{tabular}
\end{adjustbox}
\end{center}
\caption{Comparison of blind Gaussian denoising on the CBSD68 dataset. Our modified version of DnCNN for spatial denoising has comparable results with the original one. The values represent \textit{PSNR[dB]}, the higher the better. DnCNN results obtained with the provided Matlab implementation \cite{DNCNNcode}. CBSD68 available here \cite{CBSD68}.\\ \small{*Noisy images clipped in range [0,255].}}
\label{tab:tab1}
\end{table}

To further validate on real-world images, we evaluate the sRGB DND dataset \cite{DND} and submitted for evaluation. The result \cite{DNDB} are encouraging, since our trained model (called \textit{128-DnCNN Tensorflow} on the DND webpage) scored an average of $37.0343 dB$ for the PSNR and $0.9324$ for the SSIM, placing it in the first 10 positions. Interestingly, the authors of DnCNN submitted their result of a fine-tuned model, called \textit{DnCNN+}, a week later, achieving the overall highest score for SSIM, which further validates its flexibility, see Table~\ref{tab:tab2}. 

\begin{table}
\centering
\begin{adjustbox}{width=0.25\textwidth,center}
\begin{tabular}{@{}lcc@{}} \toprule
 & PSNR [dB] & SSIM \\
\midrule
Spatial-CNN & 37.0343 & 0.9324 \\
CBDNet \cite{CBDNET} & 38.0564 & 0.9421 \\
DnCNN+ \cite{DNCNN} & 37.9018 & 0.943 \\
FFDNet+ \cite{FFDNET} & 37.6107 & 0.9415 \\
BM3D \cite{CBM3D} & 34.51 & 0.8507 \\ \bottomrule
\end{tabular}
\end{adjustbox}
\caption{Results of the DND benchmark \cite{DND} on real-world noisy images. It shows that our dataset, containing different noise models, is valid for real-world image denoising, placing our Spatial-CNN in the top 10 for sRGB denoising.}
\label{tab:tab2}
\end{table}

\subsection{Temp3-CNN: Temporal CNN Training}
\begin{figure}
\begin{center}
\includegraphics[width=0.3\textwidth]{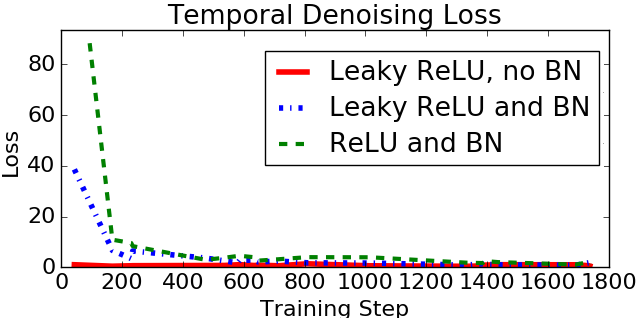}
\end{center}
\setlength{\belowcaptionskip}{-10pt}
   \caption{Evolution of the L2-Loss during the training of the Temporal-CNN. Batch Normalization (BN) does not help, adding a computation overhead without any improvement. With Leaky ReLU as activation function and with no BN, the loss starts immediately around 1 and decreases to $0.5$ after 60 epochs. Denoising without BN takes around $5\%$ less time. First 1800 steps visualized.}
\label{fig:loss}
\end{figure}

For video evaluation we need pairs of clean and noisy videos. For artificially added noise as Additive White Gaussian Noise (AWGN) or the real noise model in equation \ref{eq:3}, is easy to create such couples. However, for real-world and low-light conditions videos it is almost impossible. For this reason, this kind of video dataset, offering pairs of noisy and noise-free sequences, are not available. Therefore, we decided to proceed according to this sequence:
\smallskip
\begin{enumerate}
    \item Select 31 publicly available videos from \cite{Video}.
    \item Divide videos in sequences of 3 frames.
    \item Added either Gaussian noise with $\sigma$=[0,55] or real noise \ref{eq:3} with Ag=[0,64] and Dg=[0,32].
    \item Apply Spatial-CNN
    \item Train on pairs of spatially-denoised and clean video.
\end{enumerate}
\smallskip
We followed the same training procedure as the Spatial-CNN, even though now the network will be trained with patches of dimension $50\times50\times9$, containing three patches coming from three subsequent frames.\\
The 31 selected videos contain 8922 frames, which means 2974 sequences of three frames and a final number of patches of 300032. We ran the training for 60 epochs with a batch size of 128, Adam optimizer with learning rate of $10^{-4}$ and \textit{LeakyReLU} as activation function. It is shown LeakyReLU can outperform ReLU \cite{xu2015empirical}. However, we did not use Leaky Relu in the spatial CNN, because ReLU performed better. We present the comparison result in the supplementary material. In the final version of Temp3-CNN, Batch Normalization (BN) was not used: experiments show it slows down the training and denoising process. BN did not improve the final result in terms of PSNR. Moreover, denoising without BN requires around $5\%$ less time.
Figure \ref{fig:loss} represents the evolution of the L2-loss for the Temp3-CNN: avoiding the normalization step makes the loss starting immediately at a low value.
We trained also with Leaky ReLU, Leaky ReLU+BN and ReLU+BN and present the results in the supplementary material. 

\subsection{Exp 1: The Video Denoising CNN Architecture}
The final proposed version of \ModelName~consists in two CNNs in a pipeline, performing first spatial and then temporal denoising. To get the final architecture, we trained \ModelName~with different structures and tested it on two famous benchmarking videos and on one we personally recorded with a Blackmagic Design URSA Mini 4.6K, capable to record raw videos. The videos have various levels of Additive White Gaussian Noise (AWGN). We will answer to three critical questions.

\textbf{Q1: Is Temp3-CNN able to learn both temporal and spatial denoising?}

We compare the Spatial-CNN with the Temp3-CNN, which in this case tries to perform spatial and temporal denoising at the same time.\\
\textbf{Answer:} No, Temp3 is not enough. Referring to Table \ref{tab:q1}, we notice how using Temp3-CNN alone leads to a worse result compared to the simpler Spatial-CNN.

\begin{table}
\centering
\begin{adjustbox}{width=0.4\textwidth,center}
\begin{tabular}{@{}cccccc@{}} \toprule
 & \multicolumn{2}{c}{\textit{Foreman}} & \multicolumn{2}{c}{\textit{Tennis}} & \textit{Strijp-S} * \\ 
\midrule
 Res./Frames & \multicolumn{2}{c}{$288{\times}352$ / $300$} & \multicolumn{2}{c}{$240{\times}352$ / $150$} & $656{\times}1164$/$787$\\
\midrule
 $\sigma$ & $25$ & $55$ & $25$ & $55$ & $25$\\
\midrule
Spatial-CNN & 32.18 & 28.27 & 29.46 & 26.15 & 32.73\\
Temp3-CNN & 31.56 & 27.45 & 29.32 & 25.63 & 31.13\\
\bottomrule
\end{tabular}
\end{adjustbox}
\caption{Comparison of Spatial-CNN and Temp3-CNN over videos with different levels of AWGN. The Temp3-CNN alone can not outperform the Spatial-CNN. Results expressed in terms of \textit{PSNR[dB]}. \small{*Self-recorded Raw video converted to RGB.}}
\label{tab:q1}
\end{table}

\textbf{Q2: Ordering of spatial and temporal denoising?}
Knowing that using Temp3-CNN alone is not enough, we now have to compare different combination of spatial and temporal denoising.\\
\textbf{Answer:} looking at Table \ref{tab:q2}, we can confirm that using temporal denoising improves the result over spatial denoising, with the best performing combination as Spatial-CNN followed by Temp3-CNN.

\begin{table}
\centering
\begin{adjustbox}{width=0.4\textwidth,center}
\begin{tabular}{@{}cccccc@{}} \toprule
 & \multicolumn{2}{c}{\textit{Foreman}} & \multicolumn{2}{c}{\textit{Tennis}} & \textit{Strijp-S} \\ 
\midrule
 Res./Frames & \multicolumn{2}{c}{$288{\times}352$ / $300$} & \multicolumn{2}{c}{$240{\times}352$ / $150$} & $656{\times}1164$/$787$\\
\midrule
 $\sigma$ & $25$ & $55$ & $25$ & $55$ & $25$\\
\midrule
\thead{Spatial-CNN} & 32.18 & 28.27 & 29.46 & 26.15 & 32.73\\
\thead{Temp3-CNN \&\\Spatial-CNN} & 32.09 & 28.37 & 29.21 & 25.98 & 32.28\\
\thead{Spatial-CNN \& \\Temp3-CNN} & 33.12 & 29.56 & 30.36 & 27.18 & 34.07\\
\bottomrule
\end{tabular}
\end{adjustbox}
\caption{The combination of Spatial-CNN + Temp3-CNN is the best performing, showing consistent improvements of $\sim1dB$ over the spatial-only denoising. Results expressed in terms of \textit{PSNR[dB]}.}
\label{tab:q2}
\end{table}

~\\ \textbf{Q3: How many frames to consider?}
We compare now the introduced Temp3-CNN with Temp5-CNN, which considers a time window of 5 frames.\\
\textbf{Answer:} Results in Table \ref{tab:q3} shows that considering more frames could improve the result, but this is not guaranteed. Therefore, since using a bigger time window means more memory and time needed, we decided to use the three frames model for a better trade-off. For comparison, using the Temp5-CNN on the video \textit{Foreman} took $6.5\%$ more time than using the Temp3-CNN, $21.17s$ vs $19.85s$ on GPU.
The difference does not seem much here, but will enhance for realistic large videos.

\begin{table}
\centering
\begin{adjustbox}{width=0.4\textwidth,center}
\begin{tabular}{@{}cccccc@{}} \toprule
 & \multicolumn{2}{c}{\textit{Foreman}} & \multicolumn{2}{c}{\textit{Tennis}} & \textit{Strijp-S}\\ 
\midrule
 Res./Frames & \multicolumn{2}{c}{$288{\times}352$ / $300$} & \multicolumn{2}{c}{$240{\times}352$ / $150$} & $656{\times}1164$/$787$\\
\midrule
 $\sigma$ & $25$ & $55$ & $25$ & $55$ & $25$\\
\midrule
\thead{Spatial-CNN} & 32.18 & 28.27 & 29.46 & 26.15 & 32.73\\
\thead{Spatial-CNN \& \\Temp3-CNN} & 33.12 & 29.56 & 30.36 & 27.18 & 34.07\\
\thead{Spatial-CNN \& \\Temp5-CNN} & 33.03 & 29.87 & 30.72 & 27.70 & 33.97\\
\bottomrule
\end{tabular}
\end{adjustbox}
\caption{Comparison of architectures using 3 or 5 frames. Using a bigger time window, i.e. five frames, may slightly improve the final result or even worsen it. Hence, we decided to proceed using a 3-frames architecture. Results expressed in terms of \textit{PSNR[dB]}.}
\label{tab:q3}
\end{table}

\subsection{Exp 2: Sensitivity to Temporal Inconsistency}
\begin{figure}[h]
\centering
\begin{subfigure}{.1\textwidth}
  \centering
  \captionsetup{justification=centering}
  \includegraphics[width=1\linewidth]{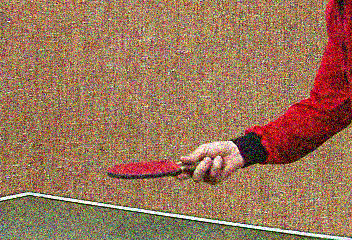}
  \caption{Noisy Frame 9}
  \label{fig:mfig1}
\end{subfigure}
\begin{subfigure}{.1\textwidth}
  \centering
  \captionsetup{justification=centering}
  \includegraphics[width=1\linewidth]{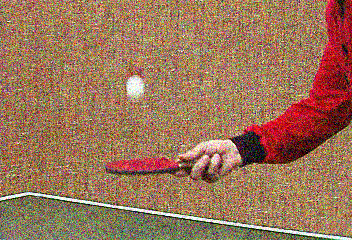}
  \caption{Noisy Frame 10}
  \label{fig:mfig2}
\end{subfigure}
\begin{subfigure}{.1\textwidth}
  \centering
  \captionsetup{justification=centering}
  \includegraphics[width=1\linewidth]{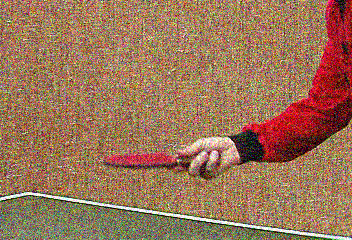}
  \caption{Noisy Frame 11}
  \label{fig:mfig3}
\end{subfigure}
\begin{subfigure}{.1\textwidth}
  \centering
  \captionsetup{justification=centering}
  \includegraphics[width=1\linewidth]{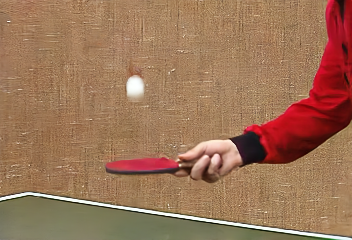}
  \caption{Denoised Frame 10}
  \label{fig:mfig4}
\end{subfigure}
\setlength{\belowcaptionskip}{-10pt}
\caption{\ModelName~achieves the same PSNR value of $27.28dB$ for frame 10 of the video \textit{Tennis} with AWGN $\sigma$=40, even if we manually cancel the white ball from the secondary frames. The network understands which part has to take into consideration and which not, i.e. the ball area.}
\label{fig:fig3}
\end{figure}

We investigate temporal consistency with a simple experiment where we remove an object from the first frame and the last frame where we denoise on the middle frame, see Figure \ref{fig:fig3}. Specifically, (i) on the video \textit{Tennis} from \cite{Video}, add Gaussian noise with standard deviation $\sigma$=40; (ii) Manually remove the white ball on the first and last frame; (iii) Denoise the middle frame. In Figure \ref{fig:fig3} we show the modified input frames and \ModelName~output. In terms of PSNR value, we got the same value for both normal and experimental case: $27.28dB$. This is an illustration that the network does what we expected: it uses part of the secondary frames and combine them with the reference, but only where the pixel content is similar enough: the ball is not removed from frame 10.

\begin{figure}[h]
\centering
\begin{subfigure}{.15\textwidth}
  \centering
  \captionsetup{justification=centering}
  \includegraphics[width=1\linewidth]{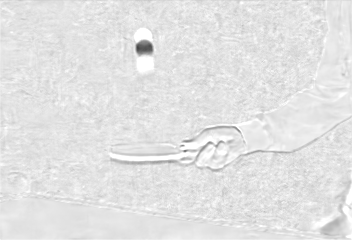}
  \caption{Filter 90}
  \label{fig:ffig3}
\end{subfigure}
\begin{subfigure}{.15\textwidth}
  \centering
  \captionsetup{justification=centering}
  \includegraphics[width=1\linewidth]{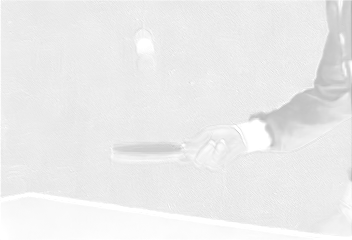}
  \caption{Filter 59}
  \label{fig:ffig2}
\end{subfigure}
\caption{Visualization of filters 59 and 90 output of Temp3-CNN first convolutional layer. We used frames number 9, 10 and 11 from the video \textit{Tennis} as input. Filter 59 highlights the reference ball and other areas with similar colors, whereas filter 90 seems to highlight mostly contours and the ball at the reference position in frame 10.}
\label{fig:filters}
\end{figure}

\textbf{Visualization of temporal filters}
To understand what our model detects, we show in Figure \ref{fig:filters} the output of two of the 128 filters in the first layer of Temp3-CNN. In Figure \ref{fig:ffig3}, we see in black the table-tennis ball of the current frame, whereas the ones in the previous and subsequent frame are in white. In Figure \ref{fig:ffig2} instead, we see how this filter highlights flat areas with similar colors and shows mostly the ball of the current frame in white. Therefore, Temp3-CNN gives different importance to similar and different areas among the three frames. This is a simple indication on how the CNN handles motion and temporal inconsistencies.

\begin{table*}
\centering
\begin{adjustbox}{width=0.7\textwidth,center}
\begin{tabular}{@{}lcccccccccccc@{}} \toprule
 & \multicolumn{3}{c}{\textit{Tennis}} & \multicolumn{3}{c}{\textit{Old Town Cross}} & \multicolumn{3}{c}{\textit{Park Run}} & \multicolumn{3}{c}{\textit{Stefan}}\\
\midrule
 Res./Frames & \multicolumn{3}{c}{$240{\times}352$ / 150} & \multicolumn{3}{c}{$720{\times}1280$ / $500$} & \multicolumn{3}{c}{$720{\times}1280$ / $504$} & \multicolumn{3}{c}{$656{\times}1164$ / $300$} \\
\midrule
 $\sigma$ & $5$ & $25$ & $40$ & $15$ & $25$ & $40$ & $15$ & $25$ & $40$ & $15$ & $25$ & $55$\\
\midrule
\ModelName & 35.51 & 29.97 & 28.00 & 32.15 & 30.91 & 29.41 & 31.04 & 28.44 & 25.97 & 32.06 & 29.23 & 24.63 \\
\ModelName-G & \textbf{37.81} & \textbf{30.36} & \textbf{28.44} & 32.39 & \textbf{31.29} & \textbf{29.97} & \textbf{31.25} & \textbf{28.72} & \textbf{26.36} & \textbf{32.37} & \textbf{29.59} & \textbf{25.06} \\
VBM4D \cite{VBM4D} & 34.64 & 29.72 & 27.49 & \textbf{32.40} & 31.21 & 29.57 & 29.99 & 27.90 & 25.84 & 29.90 & 27.87 & 23.83 \\
CBM3D \cite{CBM3D} & 27.04 & 26.37 & 25.62 & 28.19 & 27.95 & 27.35 & 24.75 & 24.46 & 23.71 & 26.19 & 25.89 & 24.18\\
DnCNN \cite{DNCNN} & 35.49 & 27.47 & 25.43 & 31.47 & 30.10 & 28.35 & 30.66 & 27.87 & 25.20 & 32.20 & 29.29 & 24.51 \\
\bottomrule
\end{tabular}
\end{adjustbox}
\caption{Comparison of \ModelName~with a video denoising algorithm, VBM4D \cite{VBM4D}, and two image denoising algorithms, DnCNN \cite{DNCNN} and CBM3D \cite{CBM3D}. \ModelName-G is the model trained specifically for blind Gaussian denoising. Test videos have different length, size and level of Additive White Gaussian Noise. \ModelName~performs better than blind denoising algorithms CBM3D, DnCNN and VBM4D, which has been used with the \textit{low complexity} setup due to our memory limitations. Best results are highlighted in bold. Original videos are publicly available here \cite{Video}. Results expressed in terms of\textit{ PSNR[dB]}.}
\label{tab:tab4}
\end{table*}

\begin{figure*}
\centering
\begin{subfigure}{.15\textwidth}
  \centering
  \includegraphics[width=.98\linewidth]{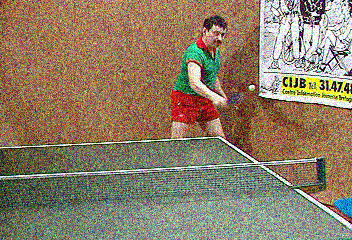}
  \captionsetup{justification=centering}
  \caption{Noisy frame\\148}
  \label{fig:147}
\end{subfigure}
\begin{subfigure}{.15\textwidth}
  \centering
  \includegraphics[width=.98\linewidth]{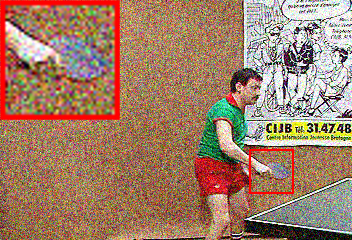}
  \captionsetup{justification=centering}
  \caption{Noisy frame\\149}
  \label{fig:sfig1}
\end{subfigure}
\begin{subfigure}{.15\textwidth}
  \includegraphics[width=.98\linewidth]{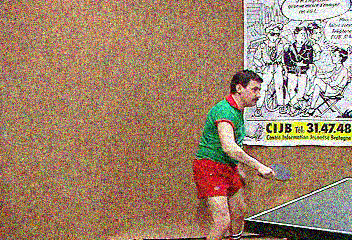}
  \caption{Noisy frame\\150}
  \label{fig:149}
\end{subfigure}%
\begin{subfigure}{.15\textwidth}
  \centering
  \includegraphics[width=.98\linewidth]{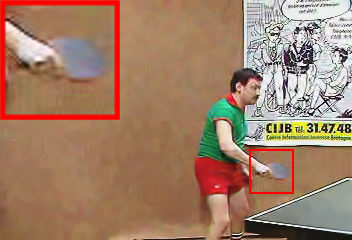}
  \captionsetup{justification=centering}
  \caption{CBM3D\cite{BM3D}\\(25.60/0.9482)}
  \label{fig:sfig2}
\end{subfigure}
\begin{subfigure}{.15\textwidth}
  \centering
  \includegraphics[width=.98\linewidth]{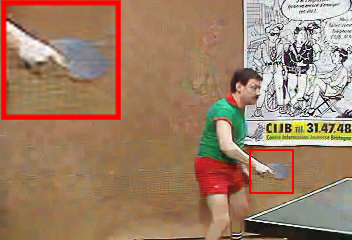}
  \captionsetup{justification=centering}
  \caption{VBM4D\cite{VBM4D}\\(24.51/0.9292)}
  \label{fig:sfig3}
\end{subfigure}
\begin{subfigure}{.15\textwidth}
  \centering
  \includegraphics[width=.98\linewidth]{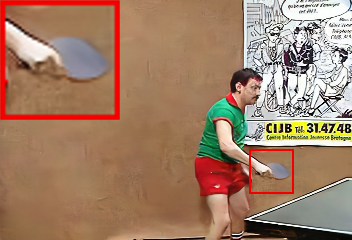}
  \captionsetup{justification=centering}
  \caption{\ModelName-G\\(27.19/0.9605)}
  \label{fig:sfig4}
\end{subfigure}
\caption{Blind video denoising comparison on \textit{Tennis} \cite{Video} corrupted with AWGN $\sigma$=40 and values clipped between [0,255]. We show the result of two competitors, VBM4D and CBM3D, which scored respectively second and third (see Table \ref{tab:tab4}) on this test video. \ModelName~performs well in challenging situations, even if the previous frame is completely different \ref{fig:147}, thanks to the temporal inconsistency detection. VBM4D suffers from the change of view, creating artifacts. Results in brackets are referred to the single frame 149 (PSNR [dB]/SSIM).}
\label{fig:fig}
\end{figure*}

\begin{table*}
\centering
\begin{adjustbox}{width=0.5\textwidth,center}
\begin{tabular}{@{}ccccccccc@{}}
\toprule
 & \multicolumn{2}{c}{Train} & Mountains & \multicolumn{3}{c}{Windmill}\\
\midrule
 Res./Frames &\multicolumn{2}{c}{ $212{\times}1091$ / 4} & $1080{\times}1920$ / 4 & \multicolumn{3}{c}{$1080{\times}1920$ / 3}\\
\midrule
Light & 50/255 & 55/255 & [55,75]/255 & 44.6 lux & 118 lux & 212 lux\\
\midrule
\ModelName & \bf{34.05} & \bf{36.96} & \bf{40.84} & \bf{32.96} & \bf{35.42} & \bf{36.69} \\
VBM4D\cite{VBM4D} & 29.10 & 33.48 & 37.34 & 26.62 & 30.69 & 32.92 \\
CBDNet\cite{CBDNET} & 30.89 & 34.56 & 39.91 & 29.56 & 34.31 & 36.22 \\
CBM3D\cite{CBM3D} & 31.27 & 34.06 & 40.20 & 29.81 & 34.06 & 35.74 \\
DnCNN\cite{DNCNN} & 24.33 & 29.87 & 32.39 & 21.73 & 25.55 & 27.87 \\
\bottomrule\end{tabular}
\end{adjustbox}
\caption{Comparison of state-of-art denoising algorithms over six low-light sequences recorded with a Bosch Autodome IP 5000 IR in raw mode, without any type of filtering activated. Every sequence is composed of 4 or 3 frames, with ground truths obtained averaging over 200 images. The \textit{Windmill} sequences has been recorded with a different light source, where we were able to measure the light intensity. Highlighted in bold our \ModelName~results, which performs well. Results expressed in terms of \textit{PSNR[dB]}.}
\label{tab:tab5}
\end{table*}

\begin{figure*}
  \centering
\begin{subfigure}{.25\textwidth}
  \centering
  \adjincludegraphics[width=\linewidth,trim={{.1\width} 0 0 0},clip]{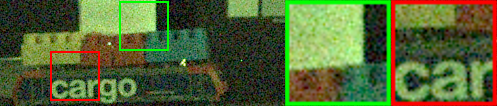}
  \caption{Noisy frame 2\\(22.54/0.4402)}
  \label{fig:0002}
\end{subfigure}
\begin{subfigure}{.25\textwidth}
  \centering
  \adjincludegraphics[width=\linewidth,trim={{.1\width} 0 0 0},clip]{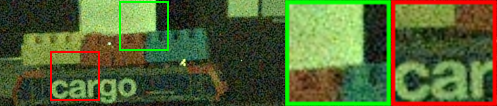}
  \captionsetup{justification=centering}
  \caption{DnCNN \cite{DNCNN}\\(24.30/0.5323)}
  \label{fig:vfig1}
\end{subfigure}
\begin{subfigure}{.25\textwidth}
  \centering
  \adjincludegraphics[width=\linewidth,trim={{.1\width} 0 0 0},clip]{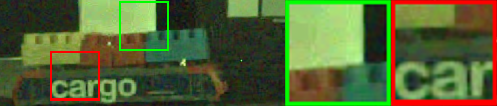}
  \captionsetup{justification=centering}
  \caption{VBM4D \cite{VBM4D}\\(29.08/0.7684)}
  \label{fig:vfig2}
\end{subfigure}
\begin{subfigure}{.25\textwidth}
  \centering
  \adjincludegraphics[width=\linewidth,trim={{.1\width} 0 0 0},clip]{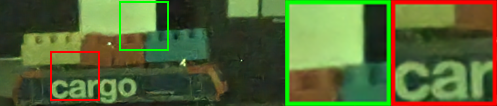}
  \captionsetup{justification=centering}
  \caption{CBDNet \cite{CBDNET}\\(30.75/0.8710)}
  \label{fig:vfig3}
\end{subfigure}
\begin{subfigure}{.25\textwidth}
  \centering
  \adjincludegraphics[width=\linewidth,trim={{.1\width} 0 0 0},clip]{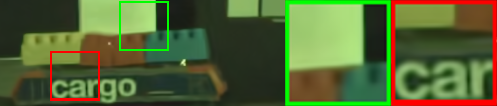}
  \captionsetup{justification=centering}
  \caption{CBM3D\cite{CBM3D}\\(31.11/0.8982)}
  \label{fig:vfig4}
\end{subfigure}
\begin{subfigure}{0.25\textwidth}
  \centering
  \adjincludegraphics[width=\linewidth,trim={{.1\width} 0 0 0},clip]{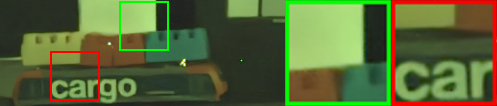}
  \captionsetup{justification=centering}
  \caption{\ModelName\\(34.14/0.9158)}
  \label{fig:vfig5}
\end{subfigure}
\caption{Detailed comparison of denoising algorithms on the low-light video \textit{Train} with light intensity at 50/255. Our \ModelName~shows good performance in this light condition, preserving edges and correctly smoothing flat areas. Results referred to frame 2, expressed in terms of (PSNR [dB]/SSIM).}
\label{fig:fig0}
\end{figure*}

\subsection{Exp 3: Evaluating Gaussian Video Denoising}
Currently, most of the video and image denoising algorithms have been developed to tackle Additive White Gaussian Noise (AWGN). We will compare \ModelName~with the state-of-art algorithm for Gaussian video denoising VBM4D \cite{VBM4D} and additionally with CBM3D \cite{CBM3D} and DnCNN \cite{DNCNN} for single frame denoising. We used the algorithms in their blind version: for VBM4D we activated the noise estimation, for CBM3D we set the sigma level to 50 and for DnCNN we use the blind model provided by the authors.
We compare two versions of \ModelName, where \ModelName-G is the model trained specifically for AWGN denoising and \ModelName~is the final model tackling multiple noise models, including low-light conditions. The videos have been stored as uncompressed png frames, added with AWGN and saved again in loss-less png format.
From the results in Table \ref{tab:tab4} we notice that VBM4D achieves superior results compared to its spatial counterpart CBM3D, which is probably due to the effectiveness of the noise estimator implemented in VBM4D. CBM3D suffers from the wrong noise std. deviation ($\sigma$) level for low noise intensities, whereas for high levels achieves comparable results.
Overall, our implemented \ModelName~in its Gaussian specific version performs better than the general blind model, even though the difference is limited. \ModelName-G scores the best results, as highlighted in bold in Table \ref{tab:tab4}, confirming our blind video denoising network as a valid approach, which achieves state-of-art results.

\subsection{Exp 4: Evaluating Low-Light Video Denoising}
Along with the low-light dataset creation, we also recorded six sequences of three or four frames each:
\begin{itemize}
      \setlength{\itemsep}{0pt}
    \item Two sequences of the same scene, with a moving toy train, in two different light intensities.
    \item A sequence of an artificial mountain landscape with increasing light intensity.
    \item Three sequences of the same scene, with a rotating toy windmill and a moving toy truck, in three different light conditions.
\end{itemize}
\smallskip
Those sequences are not part of the training set and have been recorded separately, with the same Bosch Autodome IP 5000 IR camera. In Table \ref{tab:tab5} we present the results of \ModelName~ highlighted in bold, in comparison with other state-of-art denoising algorithms on the low-light test set. We compare our method with VBM4D \cite{VBM4D}, CBM3D \cite{CBM3D}, DnCNN \cite{DNCNN} and CBDNet \cite{CBDNET}. \ModelName~outperforms the competitors, especially for the lowest light intensities. Surprisingly, the single frame denoiser CBM3D performs better than the video version VBM4D: the difference may be because CBM3D in its blind version uses $\sigma=50$, whereas VBM4D has a built-in noise level estimator, which may perform worse with a completely different noise model from the supposed Gaussian one. 

\section{Discussion}

In this paper, we presented a novel CNN architecture for Blind Video Denoising called \ModelName. We use spatial and temporal information in a feed-forward process, combining three consecutive frames to get a clean version of the middle frame. We perform temporal denoising in simple yet efficient manner, where our Temp3-CNN learns how to handle objects motion, brightness changes, and temporal inconsistencies. We do not address camera motion in videos, since the model was designed to reduce the bandwidth usage of static security cameras keeping the network as simple and efficient as possible. We define our model as \textit{Blind}, since it can tackle different noise models at the same time, without any prior knowledge nor analysis of the input signal. We created a dataset containing multiple noise models, showing how the right mix of training data can improve image denoising on real world data, such as on the DND Benchmarking Dataset \cite{DND}. We achieve state-of-art results in blind Gaussian video denoising, comparing our outcome with the competitors available in the literature. We show how it is possible, with the proper hardware, to address low-light video denoising with the use of a CNN, which would ease the tuning of new sensors and camera models. Collecting the proper training data would be the most time consuming part. However, defining an automatic framework with predefined scenes and light conditions would simplify the process, allowing to further reduce the needed time and resources. Our technique for acquiring clean and noisy low-light image pairs has proven to be effective and simple, requiring no specific exposure tuning.

\subsection{Limitations and Future Works}
The largest real-world limitations of \ModelName~is the required computational power. Even with an high-end graphic card as the Nvidia Titan X, we were able to reach a maximum speed of only $\sim3$fps on HD videos. However, most of the current cameras work with Full HD or even UHD (4K) resolutions with high frame rates. We did not try to implement \ModelName~on a mobile device supporting Tensorflow Lite, which converts the model to a lighter version more suitable for handled devices. This could be new development and challenging question to investigate on, since every week the available hardware in the market improves.



{\small
\bibliographystyle{unsrt}
\bibliography{egbib}
}

\end{document}